\pdfoutput=1

\documentclass[11pt]{article}

\usepackage[]{EMNLP2024}

\usepackage{times}
\usepackage{latexsym}

\usepackage[T1]{fontenc}

\usepackage[utf8]{inputenc}

\usepackage{microtype}

\usepackage{inconsolata}

\usepackage{graphicx}
\usepackage{amsmath}
\usepackage{multirow}
\usepackage{subcaption}

\usepackage{booktabs}
\usepackage{xspace}
\usepackage{makecell}

\usepackage{siunitx} %

\usepackage[capitalize,noabbrev]{cleveref}

\usepackage{anyfontsize}

\def\tok100{\mathcal{T}_{100}}

\newlength{\widebarargwidth}
\newlength{\widebarargheight}
\newlength{\widebarargdepth}

\newcommand{\1}{\mathbb{I}} 

\usepackage{amsmath,amsfonts,bm}

\def\eqref#1{equation~\ref{#1}}

\def\1{\bm{1}}

\DeclareMathAlphabet{\mathsfit}{\encodingdefault}{\sfdefault}{m}{sl}
\SetMathAlphabet{\mathsfit}{bold}{\encodingdefault}{\sfdefault}{bx}{n}

\usepackage{amssymb}

\usepackage{xcolor}

\usepackage{amsmath}
\usepackage{algorithm}
\usepackage{algorithmic}

\usepackage{booktabs}
\usepackage{multirow}
\usepackage{graphicx}
\usepackage{colortbl}
\usepackage{xcolor}

\usepackage{geometry}
\usepackage{array}
\usepackage{graphicx} 
\usepackage{booktabs} 
\usepackage{multirow} 
\usepackage{colortbl} 
\usepackage{xcolor}   

\title{AutoMixer: Checkpoint Artifacts as Automatic Data Mixers}

\author{
Ernie Chang$^{\spadesuit}$\textsuperscript{*} \,
Yang Li\textsuperscript{*}$^{\dagger}$ \,
Patrick Huber$^{\spadesuit}$ \,
Vish Vogeti$^{\spadesuit}$ \,
{\bf David Kant$^{\spadesuit}$} \,
{\bf Yangyang Shi$^{\spadesuit}$} \,
{\bf Vikas Chandra$^{\spadesuit}$} \\
$^\spadesuit$AI at Meta \\
$^{\dagger}$Iowa State University \\
{\tt erniecyc@meta.com, yangli1@iastate.edu}
}

\begin{document}

\maketitle

\def\thefootnote{*}\footnotetext{These authors contributed equally to this work.}\def\thefootnote{\arabic{footnote}}

\begin{abstract}

In language model training, it is desirable to equip models with capabilities from various tasks.
However, it is not clear how to directly obtain the right data mixtures for these capabilities as the relationship between data and tasks is difficult to be modeled. 
In this work, we observe that checkpoint models exhibit emerging capabilities at different points in the training trajectory.
Often, the training process saves checkpoints as artifacts that are under-utilized as a source of in-training data signals.
We identify these artifact models based on their respective capabilities on the benchmarks and leverage them as data mixers by using their aggregated first-order influence approximation over source data (See Figure~\ref{fig:overview}). 
We demonstrated on eight reasoning benchmarks that the proposed framework shows significant improvements in the pretraining setting, with performance improvements of up to 1.93\%.
Overall, this shows the potential of checkpoint models to enhance data quality and optimize data mixtures.

\end{abstract}

\section{Introduction}
\label{sec:intro}

Training effective language models involves equipping them with a diverse set of skills, which is heavily influenced by the composition of their training data. 
A primary challenge in this domain is the precise identification of task-specific data within a diverse mixture~\citep{ye2024data}—an undertaking that becomes increasingly complex as the number of tasks grows and direct domain matches (or dataset-task mapping) are absent~\citep{gadre2024datacomplm}.
This complexity is further compounded by overlapping or conflicting knowledge regions across data domains, complicating the discernment of the most relevant samples for each task~\citep{Sedinkina2019Domain}. 
Traditional methods often overlook this perspective, leading to inefficient data utilization and missed opportunities~\citep{wu-etal-2022-identifying}. 

In this work, we address this challenge as a two-fold problem: 
(1) identifying data mixtures or groups that can define effective divisions between data, and 
(2) assigning sampling weights to each of these groups to better model desirable behaviors during the training process.

\begin{figure}[t]
  \centering
\includegraphics[width=\columnwidth]{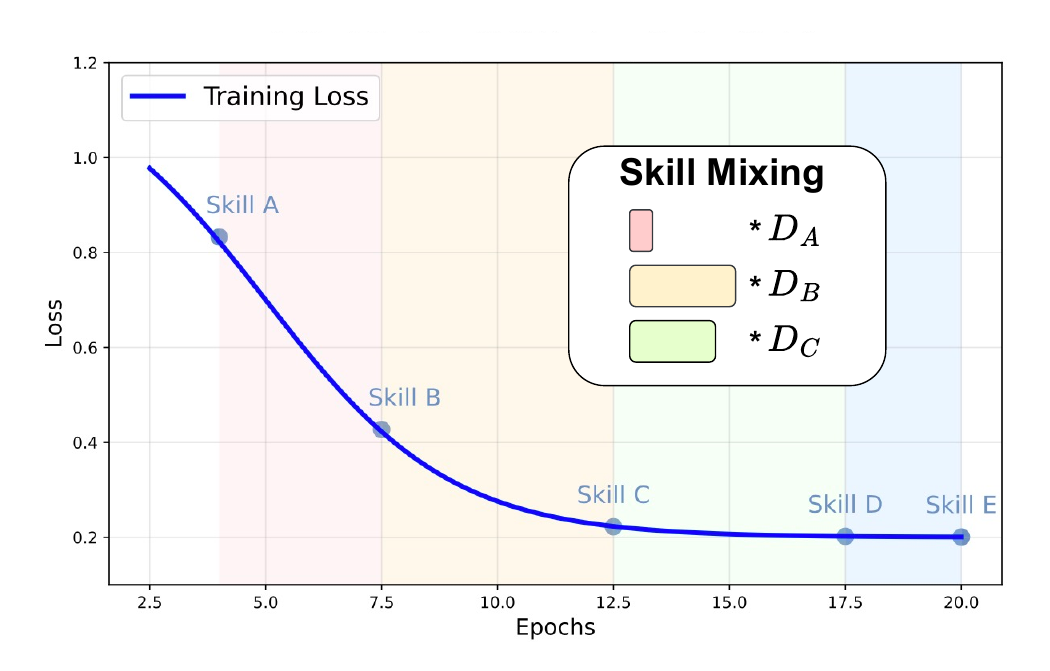}
  \caption{ Illustration of the checkpoint selection process and subsequent sampling. 
  We leverage intermediate model checkpoints to group and sample data  for targeted skill acquisition.}
  \label{fig:overview}
\end{figure}

Fundamentally, task data are often ill-defined -- it is not clear how best to assemble a dataset for cultivating a certain skillset~\citep{wei2022emergent,hu2023unlock}. 
This “chicken-and-egg” problem means that identifying the data leading to skill improvements requires a priori knowledge of which data benefits that skill. 
One could consider training a multitude of data mixtures to observe performance trends and then isolate the best data for each skill, but such brute-force approaches are computationally infeasible as models grow larger.

Therefore, the core missing piece is a direct modeling of the relationship between datasets and model parameters, because data quality cannot be reliably assessed in isolation from training~\citep{park2023trak}. 
A potential solution involves computing the influence function~\citep{hampel1974influence,halevy2009unreasonable}, which estimates the first-order ``alignment'' between training samples and specific skills. 
However, models inherently progress beyond these approximations, so data composition guided solely by step-\(t\) influence calculations may become outdated as training advances.

In this work, we propose to tackle data mixing by regrouping raw data based on observed capabilities and then assigning data loading probabilities for these groups. 
Skills acquired at one checkpoint may not persist throughout the entire process, making it difficult to identify a single training step that captures all optimal capabilities. 
Hence, we simulate training runs with proxy models and trace checkpoint artifacts that align with target tasks. 
We then estimate the sample influence~\citep{kwondatainf,yeh2022first} for each checkpoint; these influence scores are consolidated to guide both the grouping of task-relevant data and the determination of sampling weights, ultimately maximizing task influence across all data.

Our contributions are as follows: 
(1) we propose the AutoMixer framework that identifies task-specific checkpoints from simulation runs, which can be used as effective data samplers to boost model performance on desirable tasks, enabling both the identification of task data mixtures and their respective importance weights; 
(2) we show through extensive analysis that proxy models can serve as effective data samplers and mixers using only simulation runs, which allows the reuse of these proxy data mixers across different tasks and training scenarios.

\section{Background}
\label{sec:background}

\begin{figure*}[ht]
  \centering
  \includegraphics[width=\textwidth]{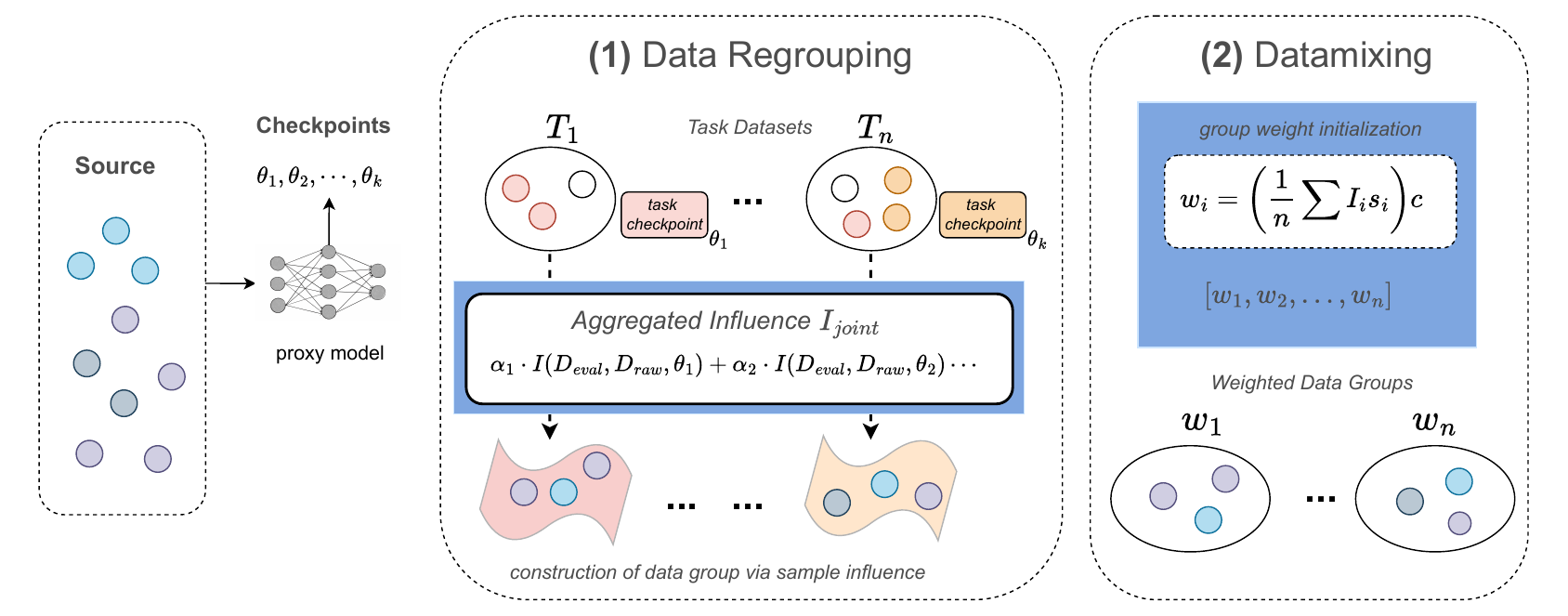}
  \caption{ Illustration of the AutoMixer framework: Each data sample from ungrouped raw pretraining sources is assigned an influence score. These scores guide the regrouping of incoming data into task-specific datasets by 
  (1) splitting raw data into groups based on task checkpoints, and
  (2) Determine sampling weights by aggregating influence scores across checkpoints.}
  \label{fig:main_concept}
\end{figure*}

Training large language models is fundamentally shaped by both the breadth and quality of their data~\citep{ye2024data}. 
A major challenge arises from identifying task-specific examples within extensive and heterogeneous corpora, particularly when direct mappings between data domains and target tasks are absent or ambiguous~\citep{Sedinkina2019Domain,wu-etal-2022-identifying}. 
Beyond this complexity, overlapping knowledge regions often blur domain boundaries, further complicating the extraction of samples most conducive to skill development~\citep{Sedinkina2019Domain,britz2017effective}. 
Traditional data-mixing techniques frequently overlook these overlaps, leading to suboptimal data usage and 
task generalizabilities~\cite{conneau2019cross,li2023word,blitzer2007domain,lee2022deduplicating,wang2020multi}.

A second layer of complexity stems from the dynamic nature of skill acquisition in language models. 
Model capabilities can surface and evolve at various points during training, often following non-monotonic trajectories~\citep{wei2022emergent,hu2023unlock}. 
This non-monotonicity gives rise to a “chicken-and-egg” dilemma: determining which data samples facilitate a particular skill is difficult until the model has already begun to exhibit that skill. 
Although brute-force methods—such as training multiple data mixtures and rigorously testing their impact on downstream tasks—could theoretically illuminate these relationships, they are computationally infeasible at scale.

Influence-based techniques offer a more principled approach to this problem. 
By examining how individual training samples affect model predictions~\citep{hampel1974influence,halevy2009unreasonable}, influence functions pinpoint data regions that are especially valuable for specific tasks. 
However, standard influence estimation and its approximations~\citep{kwondatainf,yeh2022first} often rely on a single checkpoint (commonly the final model state), thereby neglecting how early-stage knowledge may influence performance in subsequent stages~\citep{park2023trak}. 
As a result, purely first-order or single-step influence evaluations may fail to capture the full evolution of a model’s skill trajectory.

Addressing these gaps requires a framework that:
\begin{enumerate}
    \item Subdivides large, heterogeneous data sources into groups aligned with emerging model competencies, and
    \item Adjusting sampling weights for these data groups for task-aware data loading.
\end{enumerate}

Such an approach would exploit the strengths displayed at different checkpoints, rather than treating the model as static. 
Incorporating multi-checkpoint influence measurements enables adaptive data curation that aligns with the model’s ever-shifting learning needs, ultimately leading to more efficient and effective skill acquisition.

\section{The AutoMixer Framework}

In this work, we assume the presence of raw data of various sources (e.g. common crawl snapshots) comprising \(n\) samples distributed across \(m\) task data, with a token budget \(T\). 
Each sample \(x_i\), where \(i=1, \ldots, n\), consists of \(s_i\) tokens. 
The main idea behind \emph{AutoMixer} is to decide for each task whether a piece of data should be part of the training process. If the data is chosen, it becomes part of a special collection tailored for that task. This approach ensures that each task gets the most relevant data for its training needs.

Concretely, this process of \emph{data regrouping} is achieved by utilizing performance-based checkpoints, in which AutoMixer identifies the most effective stopping points during training (See Figure~\ref{fig:main_concept}). 
The idea is that when a model achieves peak performance on task \(j\) at a particular training step, the checkpoint can be employed as an effective sampler to compute influence estimation~\cite{koh2019influence,koh2019accuracy,ting2018optimal} for task  \(j\).
Empirically, we obtain the checkpoints from simulation runs, and leverage them to identify task-optimal datasets in a two-step process (See Figure~\ref{fig:main_concept}).
The aggregated task influences can then be used to determine the weights over each group, which dictates the sampling probability of groups during language model pretraining.\footnote{Suppose a data group has a weight $w_g = 0.2$, and all weights are normalized to $1$, this means, on average, $20$\% of the tokens sampled per batch will come from this group.}

\begin{table*}[t]

    \centering

    \resizebox{\textwidth}{!}{%

    \begin{tabular}{c c c c c c c c c}

    \toprule

    \multirow{2}{*}{\textsc{Model}} & \multicolumn{8}{c}{\textsc{Task}} \\ 

    & \textsc{ARC-Easy} & \textsc{ARC-Hard} & \textsc{BOOLQ} & \textsc{PIQA} & \textsc{SIQA}  & \textsc{HELLASWAG} & \textsc{OBQA} & \textsc{WINOGRANDE} \\ 

    \midrule

    \textsc{25M} & \cellcolor{cyan!20}81\% & \cellcolor{orange!20}76\% & \cellcolor{yellow!20}10\% & \cellcolor{green!20}100\% & \cellcolor{pink!20}56\% & \cellcolor{green!20}100\% & \cellcolor{orange!20}76\% & \cellcolor{cyan!20}81\% \\

    \textsc{50M} & \cellcolor{red!20}85\% & \cellcolor{blue!20}45\% & \cellcolor{purple!20}65\% & \cellcolor{violet!20}80\% & \cellcolor{teal!20}60\% & \cellcolor{green!20}100\% & \cellcolor{green!20}100\% & \cellcolor{gray!20}25\% \\

    \textsc{75M} & \cellcolor{brown!20}70\% & \cellcolor{brown!20}70\% & \cellcolor{yellow!20}5\% & \cellcolor{red!20}85\% & \cellcolor{teal!20}60\% & \cellcolor{green!20}100\% & \cellcolor{magenta!20}95\% & \cellcolor{green!20}100\% \\

    \textsc{350M} & \cellcolor{magenta!20}95\% & \cellcolor{brown!20}70\% & \cellcolor{lime!20}40\% & \cellcolor{green!20}100\% & \cellcolor{violet!20}80\% & \cellcolor{green!20}100\% & \cellcolor{magenta!20}95\% & \cellcolor{green!20}100\% \\

    \textsc{500M} & \cellcolor{green!20}100\% & \cellcolor{magenta!20}95\% & \cellcolor{olive!20}75\% & \cellcolor{red!20}85\% & \cellcolor{orange!20}90\% & \cellcolor{green!20}100\% & \cellcolor{green!20}100\% & \cellcolor{red!20}85\% \\

    \textsc{1.5B} & \cellcolor{red!20}85\% & \cellcolor{red!20}85\% & \cellcolor{red!20}85\% & \cellcolor{magenta!20}95\% & \cellcolor{orange!20}90\% & \cellcolor{magenta!20}95\% & \cellcolor{magenta!20}95\% & \cellcolor{magenta!20}95\% \\

    \bottomrule

    \end{tabular}%

    }

    \caption{ 
    Checkpoint progression ratio for various tasks across different model sizes for a total of 100K steps:
    For instance, if the checkpoint that performs the best at a task is stored at the $5000$th step, it is then recorded on the table as 5\%.
    Checkpoints for each model size are collected from one simulation run by training the language model on FineWeb data~\cite{lozhkov2024fineweb-edu} with 100K steps. 
    Here we set the tasks to be the considered benchmarks: ARC-easy, ARC-challenge~\cite{clark2018arc}, BoolQ~\cite{clark2019boolq}, PIQA~\cite{bisk2020piqa}, SIQA~\cite{sap2019siqa}, HellaSwag~\cite{zellers2019hellaswag}, OBQA~\cite{mihaylov2018obqa}, and WinoGrande~\cite{sakaguchi2021winogrande}.}
    \label{tab:checkpoints}
\end{table*}

\begin{enumerate}
    \item \textbf{Data Regrouping}: First, raw data are regrouped based on the sampled checkpoints (\(\theta_1, \dots, \theta_k\)). These \(k\) checkpoints are selected from a single simulation run, each corresponding to the best-performing checkpoint for one of the \(m\) tasks, where \(k \leq m\) (see Table~\ref{tab:checkpoints}).\footnote{In the case where $m$ becomes large, we can perform unsupervised clustering to keep it manageable.} 
    To quantify the alignment score with the task data, influence scores are obtained from simulation runs with proxy language models, which are smaller and faster to compute sample influence. 
    \item \textbf{Datamix Reweighting}: Next, to assign sampling weights to each data group, the per-sample influence is aggregated with sample token counts in a reweighting process.
    This maximizes the influence across all tasks but also ensures token count constraints are met.
\end{enumerate}

We delve into these two steps as follows.

\section{Data Regrouping via Sample Influence}
\begin{figure}[t]
  \centering
  \includegraphics[width=1.0\columnwidth]{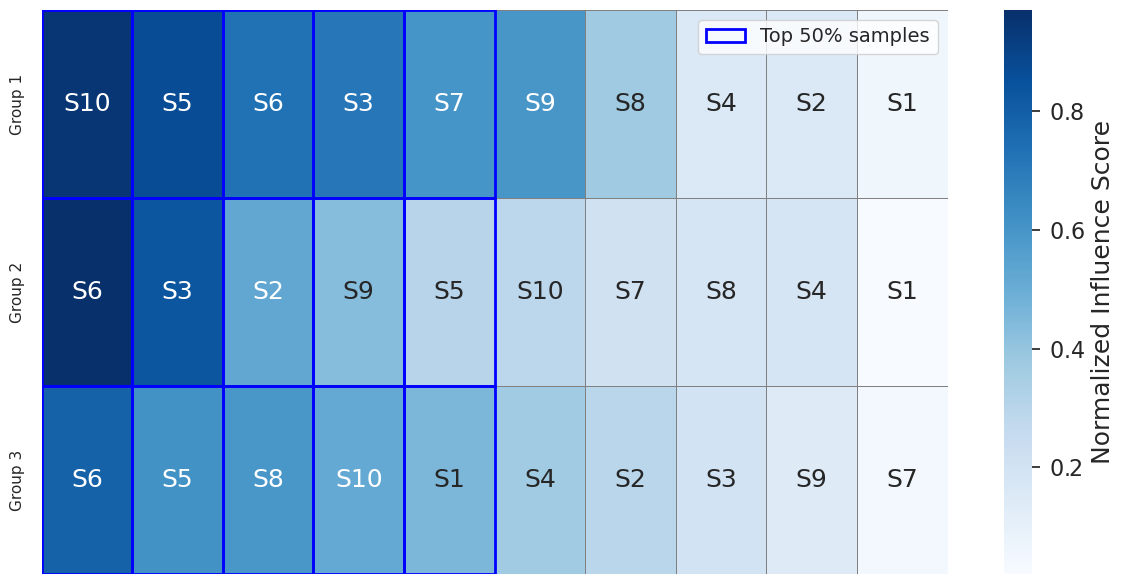}
  \caption{ 
  \textbf{Depiction of the Data Regrouping Process:}
Within each group, samples (w/ indices) are sorted based on their joint influence scores across all tasks. This sorting results in different sample orderings between groups. 
The final step involves selecting the top \(K\%\) of samples from each group to form a data group that fulfills the token budget.
Although there are duplicates across the table, we found that repeated samples contain high-value tokens that are beneficial for repeated exposure during training.
  }
  \label{fig:grouping}
\end{figure}

Data regrouping enhances language model pretraining by reorganizing raw data into task-specific groups, each defined by a distinct model checkpoint sampled from a simulation run.
During the simulation training, a collection of checkpoints is obtained. 
We record the performance of each checkpoint across all tasks, allowing us to create a benchmark table summarizing task performances, as shown in Table~\ref{tab:checkpoints}\footnote{Same checkpoint numbers are indicated with same colors.}.
We then select the top $k$ checkpoints that perform best across all $m$ tasks, where $k$ is less than or equal to $m$.
The final step involves regrouping data samples based on their utility scores, defined as influence scores~\cite{koh2019influence}. In our experiments, influence scores \(\mathcal{I}(x_i)\) are calculated for all samples using proxy models with 75M and 350M parameters.
Within each of the \(m\) task-aligned groups, samples are sorted by their utility scores, retaining the top 50\% of samples for each checkpoint sampler (See Figure~\ref{fig:grouping}).
For each sample \(x_i\), we aggregate contributions from all checkpoints:
\[
\mathcal{I}_{\text{joint}}(x_i) = \sum_{j=1}^k \alpha_j \cdot \mathcal{I}(x_i; \theta_j),
\]
where \(\alpha_j\) is the blending factor for checkpoint \(\theta_j\), determined by task performance and the checkpoint number, which we will in the next paragraph.

For simulation runs, the token budget is set at 100,000 steps with a batch size of 8, across 4 nodes, and a sequence length of 2048, totaling 6.4 billion tokens.
This setup is designed to limit the cost of simulation runs by keeping the total computing budget in a reasonably range;
and also for the fact that the proxy models tend to converge early in training.
In cases where data groups exceed this token budget, each group is subsampled in proportion to its aggregate influence scores, ensuring optimal alignment between tokens and tasks while adhering to computational constraints.

\paragraph{\emph{Blending Factor} via Task Acquisition Speed.}

Moreover, Table~\ref{tab:checkpoints} also implies the speed of convergence of different tasks across model sizes. Namely, HellaSwag seems to take longer for the model to get good at. 
While it might be tempting to say that this is due to the difficulty of that particular task, we leave that discussion for future works. However, it does provide a decent measure of how long it takes to learn each task in the training runs.

In practice, we utilize each checkpoint to assess the influence of individual samples, while simultaneously gathering task-specific signals across all samples. To achieve this, we propose using the checkpoint steps associated with each task as weights when calculating a sample's influence for each checkpoint.
The blending factor (\(\alpha_j\)) for each checkpoint is determined by normalizing the checkpoint step numbers. Let \( s_j \) be the step number for checkpoint \( j \), and \( s_{\max} \) be the largest step number among all checkpoints. The normalized step number for checkpoint \( j \) is:

\[
\tilde{s}_j = \frac{s_j}{s_{\max}}
\]

The blending factor \(\alpha_j\) is then:

\[
\alpha_j = \frac{\tilde{s}_j}{\sum_{i=1}^k \tilde{s}_i}
\]

This ensures that the blending factors sum to 1 where the \emph{blending factor} $\alpha_i$ allows us to integrate influence scores across different tasks for each sample effectively.

Below, we formalize the influence computation framework and its implementation.

\paragraph{Formulating Sample Influence.}
Consider a pretraining dataset \(\mathcal{D} = \{(x_i, y_i)\}_{i=1}^n\) for a next-token prediction objective, where each input \(x_i\) and label \(y_i\) is a sequence of tokens, with \(y_i\) obtained by shifting \(x_i\) one token to the left.
The empirical risk minimizer \(\theta^*\) is obtained through:
\[
\theta^* := \arg\min_{\theta \in \Theta} \frac{1}{n} \sum_{i=1}^n \ell(y_{i}, f_\theta(x_i))
\]
where \(\ell\) denotes cross-entropy loss. To measure the influence of sample \((x_k, y_{k})\), we analyze parameter shifts under \(\varepsilon\)-weighted risk minimization:
\begin{align*}
\theta^{(k)}(\varepsilon) := \arg\min_{\theta \in \Theta} \Bigg( & \frac{1}{n} \sum_{i=1}^n \ell(y_{i}, f_\theta(x_i)) \\
& + \varepsilon \ell(y_{k}, f_\theta(x_k)) \Bigg)
\end{align*}

The influence of \((x_k, y_k)\) on the model's performance, reflected in the model’s predictive capability on the validation set \(\mathcal{D}^\mathrm{(val)} = \{(x_i^\mathrm{(val)}, y_i^\mathrm{(val)})\}_{i=1}^{q}\), is quantified as:
\[
\begin{aligned}
\mathcal{I}(x_k, y_k)
&= \left.
    \frac{\mathrm{d}}{\mathrm{d}\varepsilon}
    \Bigl(
      \frac{1}{q}
      \sum_{j=1}^{q}
      \ell\bigl(y_j^{(\mathrm{val})}, f_{\theta}(x_j^{(\mathrm{val)}})\bigr)
    \Bigr)
  \right|_{\varepsilon=0}\\
&\quad
  = -\Bigl(
    \frac{1}{q}
    \sum_{j=1}^{q}
    \nabla_{\theta} \ell \bigl(y_j^{(\mathrm{val})}, f_{\theta}(x_j^{(\mathrm{val)}})\bigr)
  \Bigr)\\
&\quad
  \times H^{-1}(\theta)\,\nabla_{\theta} \ell \bigl(y_k, f_{\theta}(x_k)\bigr)
  \Big|_{\theta=\theta^*}
\end{aligned}
\]
where \(H(\theta)\) is the Hessian of the empirical loss. Direct computation is prohibitive for large models, necessitating approximations.

\paragraph{Efficient Influence Approximation.}
We adopt DataInf~\cite{kwondatainf} to bypass explicit Hessian inversion. For layer \(l\) in the Transformer model:

1. Compute validation gradients averaged over \(q\) validation samples:
\[
v_l = \frac{1}{q} \sum_{j=1}^{q} \nabla_{\theta_l} \ell(y_j^{\mathrm{(val)}}, f_\theta(x_j^{\mathrm{(val)}}))
\]

2. Compute the layer-specific regularization parameter \(\lambda_l\):
\[
\lambda_l = 0.1 \times (n d_l)^{-1} \sum_{i=1}^n \|\nabla_{\theta_l} \ell(y_i, f_\theta(x_i))\|_2^2
\]
where \(d_l\) denotes the layer dimension, and \(n\) denotes the number of training samples.

3. Update running sum \(r_l\) across training samples for approximating the inverse Hessian-vector product:
\[
\begin{aligned}
c_{li} &= \frac{v_l^\top \nabla_{\theta_l} \ell(y_i, f_\theta(x_i))}{\lambda_l + \|\nabla_{\theta_l} \ell(y_i, f_\theta(x_i))\|_2^2}, \\
r_l &\gets r_l + \frac{v_l - c_{li} \nabla_{\theta_l} \ell(y_i, f_\theta(x_i))}{n \lambda_l}
\end{aligned}
\] 

4. Aggregate layer contributions to compute final influence:
\[
\mathcal{I}(x_k, y_k) \approx -\sum_{l \in \{1, L\}} r_l^\top \nabla_{\theta_l} \ell(y_k, f_\theta(x_k))
\]

\paragraph{Discriminative Layer Selection.}
In practice, we save computes by avoiding matrix multiplication across all layers of the model. 
Instead, we resort to computing influence scores with only the embedding and last layer in order to enhance influence score discriminability.
Thus, we focus on gradients from the first (embedding) and last (output) Transformer layers:
we postulate that this dual-layer strategy mitigates the \emph{cancellation effect} by using only the last layer~\cite{yeh2022first} prevalent in intermediate layers, where shared processing logic obscures sample-specific influences. 
By isolating gradient signals from these critical layers, AutoMixer obtains more reliable influence estimates for regrouping decisions.


\section{Datamix Reweighting}

Regrouped datasets are presented to the model during pretraining at varying probabilities. 
This means that data group $i$ will be sampled differently from another group $j$, if \(i \neq j\).
The datamix reweighting aims to determine the probabilities to sample from each group, which will translate to the proportion of tokens in each batch of data during training.

In data regrouping, we aggregate influence scores across selected checkpoints to optimize data sampling weights via joint influences.  
A joint influence score for each sample $(x_i)$ combines its contributions to all tasks and is computed in the data regrouping stage in the previous section as $\mathcal{I}_{\text{joint}}(x_i)$. 
Now, the question remains as to how to define the sampling weight $w_g$ for group $g$.
To reflect the impact of each data group on the overall performance of all tasks, we define the group influence density $\rho_g$ for each data group $g$ as:
\[
\rho_g = \frac{1}{T_g} \sum_{x_i \in g} \mathcal{I}_{\text{joint}}(x_i) \cdot s_i,
\]  
where $s_i$ is the token count of sample $x_i$, and $T_g$ is the token count of group $g$.  

We determine each group's sampling weight based on its influence density: 
\[
w_g = \frac{\rho_g}{\sum_{g'} \rho_{g'}},
\]  
where $w_{g}\in \left( 0, 1 \right)$. When sampling tokens from each group to form the pretraining dataset, we ensure that the proportion of tokens selected from group $g$ is $w_g$. This approach optimizes overall performance across all tasks while ensuring fairness among them. 

During the pretraining dataset construction, we track the total token count of the selected samples to ensure the total training token budget $T$ is not exceeded. In scenarios where we want to limit the token budget for a specific group, we sum the token counts of that group’s selected samples. Once the group’s limit is reached, we stop sampling from it and proportionally increase the sampling weights for the remaining groups. This ensures the token budget of the group is respected.




\section{Experimental Setup}

\paragraph{Dataset.} 
The experiments employ the FineWeb-Edu dataset~\cite{lozhkov2024fineweb-edu}\footnote{Open Data Commons Attribution License (ODC-By) v1.0}, a specialized educational corpus derived from FineWeb through quality-based curation. Two versions are available: a foundational 1.3 trillion token collection and an expanded 5.4 trillion token iteration (FineWeb-Edu-score-2). The dataset ensures educational relevance through a classifier trained on Llama3-70B-Instruct-generated synthetic annotations, which selects pedagogically valuable content.

\begin{table*}[t]
    \centering
    \resizebox{\textwidth}{!}{
    \begin{tabular}{lccccccccc}
    \toprule
    \multirow{2}{*}{\textsc{Approach}}
    & \multicolumn{9}{c}{\textsc{Improvement over Uniform Sampling (Accuracy \%)}} \\
    \cmidrule(lr){2-10}
    & \textsc{ARC-Easy} & \textsc{ARC-Hard} & \textsc{BOOLQ}
    & \textsc{PIQA} & \textsc{SIQA} & \textsc{HELLASWAG} & \textsc{OBQA}
    & \textsc{WINOGRANDE} & \textsc{Avg.} \\
    \midrule
    \multicolumn{10}{c}{\textbf{350M Parameters}} \\
    \midrule
    \textsc{PPL Sampling} &
    0.35 $\pm$ 0.03 &       
    0.60 $\pm$ 0.05 &       
    0.44 $\pm$ 0.03 &       
    0.70 $\pm$ 0.04 &       
   -0.10 $\pm$ 0.05 &       
    0.55 $\pm$ 0.03 &       
    0.40 $\pm$ 0.04 &       
    0.90 $\pm$ 0.06 &       
    0.66 $\pm$ 0.03 \\      

    \textsc{n-gram sampling} &
    0.74 $\pm$ 0.06 &      
    \textbf{1.22} $\pm$ 0.04 & 
    0.79 $\pm$ 0.07 &      
    1.03 $\pm$ 0.05 &      
    1.09 $\pm$ 0.06 &      
    0.62 $\pm$ 0.03 &      
    1.16 $\pm$ 0.09 &      
    0.85 $\pm$ 0.04 &      
    0.60 $\pm$ 0.05 \\

    \textsc{AutoMixer-75M} &
   -0.15 $\pm$ 0.07 &
    0.12 $\pm$ 0.04 &
   -0.14 $\pm$ 0.05 &
    0.01 $\pm$ 0.08 &
   -0.10 $\pm$ 0.03 &
    0.05 $\pm$ 0.09 &
   -0.03 $\pm$ 0.06 &
   -0.05 $\pm$ 0.04 &
   -0.04 $\pm$ 0.05 \\    

    \textsc{AutoMixer-350M} &
    \textbf{2.23} $\pm$ 0.08 & 
    0.55 $\pm$ 0.06 &         
    \textbf{2.16} $\pm$ 0.09 &
    \textbf{2.05} $\pm$ 0.07 &
    \textbf{2.12} $\pm$ 0.10 &
    \textbf{2.33} $\pm$ 0.06 &
    \textbf{2.01} $\pm$ 0.08 &
    \textbf{2.14} $\pm$ 0.05 &
    \textbf{1.93} $\pm$ 0.07 \\
    \midrule
    \multicolumn{10}{c}{\textbf{1.5B Parameters}} \\
    \midrule
    \textsc{PPL Sampling} &
    0.20 $\pm$ 0.07 &
    0.52 $\pm$ 0.03 &
    0.32 $\pm$ 0.06 &
    0.40 $\pm$ 0.02 &
    0.07 $\pm$ 0.05 &
    0.18 $\pm$ 0.08 &
    0.75 $\pm$ 0.03 &
    0.68 $\pm$ 0.06 &
    0.48 $\pm$ 0.04 \\

    \textsc{n-gram sampling} &
    0.88 $\pm$ 0.06 &  
    \textbf{0.82} $\pm$ 0.04 &  
    1.02 $\pm$ 0.07 &  
    0.58 $\pm$ 0.05 &  
    0.45 $\pm$ 0.09 &  
    1.22 $\pm$ 0.03 &  
    0.54 $\pm$ 0.05 &  
    0.90 $\pm$ 0.08 &  
    0.79 $\pm$ 0.06 \\

    \textsc{AutoMixer-75M} &
   -0.06 $\pm$ 0.04 &
    0.01 $\pm$ 0.06 &
   -0.04 $\pm$ 0.05 &
   -0.02 $\pm$ 0.07 &
   -0.05 $\pm$ 0.08 &
    0.02 $\pm$ 0.06 &
    0.01 $\pm$ 0.04 &
   -0.03 $\pm$ 0.05 &
   -0.02 $\pm$ 0.06 \\

    \textsc{AutoMixer-350M} &
    \textbf{1.26} $\pm$ 0.05 &  
    0.39 $\pm$ 0.09 &        
    \textbf{1.35} $\pm$ 0.06 &
    \textbf{1.22} $\pm$ 0.08 &
    \textbf{1.38} $\pm$ 0.06 &
    \textbf{1.45} $\pm$ 0.04 &  
    \textbf{1.33} $\pm$ 0.07 &  
    \textbf{1.41} $\pm$ 0.09 &  
    \textbf{1.22} $\pm$ 0.05 \\
    \midrule
    \multicolumn{10}{c}{\textbf{3B Parameters}} \\
    \midrule
    \textsc{PPL Sampling} &
    0.18 $\pm$ 0.06 &
    0.32 $\pm$ 0.04 &
    0.10 $\pm$ 0.07 &
    0.42 $\pm$ 0.05 &
    0.27 $\pm$ 0.08 &
    0.34 $\pm$ 0.03 &
    0.50 $\pm$ 0.05 &
    0.15 $\pm$ 0.09 &
    0.20 $\pm$ 0.06 \\

    \textsc{n-gram sampling} &
    0.82 $\pm$ 0.05 &
    \textbf{0.72} $\pm$ 0.07 &  
    0.57 $\pm$ 0.06 &
    0.54 $\pm$ 0.04 &
    0.64 $\pm$ 0.08 &
    0.42 $\pm$ 0.05 &
    0.95 $\pm$ 0.09 &
    0.84 $\pm$ 0.06 &
    0.50 $\pm$ 0.05 \\

    \textsc{AutoMixer-75M} &
   -0.02 $\pm$ 0.06 &
    0.01 $\pm$ 0.05 &
   -0.03 $\pm$ 0.08 &
    0.00 $\pm$ 0.04 &
   -0.02 $\pm$ 0.07 &
   -0.01 $\pm$ 0.06 &
   -0.04 $\pm$ 0.03 &
    0.02 $\pm$ 0.08 &
   -0.01 $\pm$ 0.06 \\    

    \textsc{AutoMixer-350M} &
    \textbf{1.09} $\pm$ 0.04 &  
    0.34 $\pm$ 0.08 &        
    \textbf{1.12} $\pm$ 0.06 &
    \textbf{1.06} $\pm$ 0.07 &
    \textbf{1.14} $\pm$ 0.05 &
    \textbf{1.27} $\pm$ 0.09 &
    \textbf{1.23} $\pm$ 0.08 &
    \textbf{1.18} $\pm$ 0.05 &
    \textbf{1.05} $\pm$ 0.07 \\
    \bottomrule
    \end{tabular}}
    \caption{
    Improvements over the uniform baseline (accuracy \%), each with two decimal places and averaged over 2 runs.
    Negative values (e.g.\ $-0.10$) indicate worse performance than uniform for those specific tasks.
    Bolded entries represent the highest improvement in each \emph{column}.
    Overall, \textsc{AutoMixer-350M} yields the largest gains across most tasks, while other methods occasionally underperform vs.\ uniform.
    }
    \label{tab:improvement_results}
\end{table*}

\paragraph{Experimental Details.} 
We implement decoder-only transformers following the Llama-3 architecture~\cite{dubey2024llama}, pretrained with causal language modeling objectives. 
We conduct two  training runs per configuration with distinct random initializations. Our model scale analysis spans four parameter counts (350M, 1.5B, 3B) to systematically investigate size-performance relationships. 
All training occurs on a 32-GPU cluster (4 nodes, 8xH100 GPUs/node) using consistent hyperparameters across configurations.

Data selection employs influence scores calculated via the 350M proxy model, with ablation performed with the smaller 75M scale in order to understand the balance computational tractability and model capability\footnote{The estimated costs for 350M and 75M simulation runs are \$81.60 and \$79.89, respectively.}. 
Checkpoint evaluation requires $\sim$120 hours on 100 GPUs, with subsequent simulation runs completing within 48 hours. 

\paragraph{Evaluation Tasks.} 
We assess zero-shot performance on eight common-sense reasoning benchmarks: ARC-easy, ARC-challenge~\cite{clark2018arc}, BoolQ~\cite{clark2019boolq}, PIQA~\cite{bisk2020piqa}, SIQA~\cite{sap2019siqa}, HellaSwag~\cite{zellers2019hellaswag}, OBQA~\cite{mihaylov2018obqa}, and WinoGrande~\cite{sakaguchi2021winogrande}. These tasks serve dual purposes: guiding influence score computation during data selection and providing final performance metrics. This closed-loop design ensures alignment between training dynamics and evaluation objectives.

\paragraph{Benchmark Comparisons.} 
AutoMixer employ several different proxy model sizes of 75M and 350M, which we denote as \emph{AutoMixer-75M} and \emph{AutoMixer-350M} respectively. 
We evaluate against several baseline strategies:
\begin{enumerate}
    \item \textbf{Uniform Sampling}: Draws data uniformly from FineWeb without prioritization, constrained only by a fixed token budget. This baseline measures inherent dataset quality.
    \item \textbf{PPL Sampling}: Here we adopt the commonly used sampling technique~\cite{wenzek-etal-2020-ccnet} where we estimate the sample utility based on the sample sequence cross-entropy loss, or perplexity, which is used in place of the influence scores, while keeping the framework algorithm constant, where lower perplexity samples are better. 
    \item \textbf{N-gram Sampling}: Moreover, we also compared  a recent approach in \citet{chang2024target} (\emph{n-gram sampling}) that shares similar settings, where target evaluations are utilized to sample pretraining data using n-gram-based techniques~\cite{xu2020ngram}. This serves as a robust test to determine if simpler methods can achieve comparable results.
\end{enumerate}

\begin{table}[t]
    \centering
    \resizebox{\columnwidth}{!}{
    \begin{tabular}{lc}
    \toprule
    \emph{Checkpoint Strategy} & \emph{Avg. Improvement over Uniform (\%)} \\ 
    \midrule
    \emph{Last Checkpoint} & 0.7 $\pm$ 0.4 \\
    \emph{10 Checkpoints} & 0.8 $\pm$ 0.3 \\
    \emph{AutoMixer-350M} & \textbf{1.22} $\pm$ 0.05 \\
    \bottomrule
    \end{tabular}}
    \caption{ Average improvements over a uniform sampling baseline, aggregated across multiple benchmarks for 1.5B.}
    \label{tab:checkpoint_sampling}
\end{table}

\section{Main Results}

Table~\ref{tab:improvement_results} shows how different sampling strategies improve over a uniform baseline across four model scales (350M, 1.5B, and 3B parameters). 
These results support our main claim that precise identification and reweighting of task-relevant data can yield substantial performance gains. 
Specifically, \emph{AutoMixer-350M} achieves the highest overall improvements in most settings, confirming that the alignment between a proxy model and the final target model is pivotal for learning task-specific data representations. 
We also observed that it lags behind \emph{n-gram sampling} at times while \emph{AutoMixer-75M} performs poorly, suggesting the proxy model size do play a huge role in the framework effectiveness. 
Further, \emph{ppl sampling} performs poorly on the framework setting, suggesting that checkpoints cannot be used off-the-shelf in the way it intends to.
We also observe the largest gains seem to be for the 350M model, which is the proxy model size; 
which makes sense from the perspective that influence scores are computed with the same number of parameters. 

\begin{figure}[t]
  \centering
  \includegraphics[width=1.0\columnwidth]{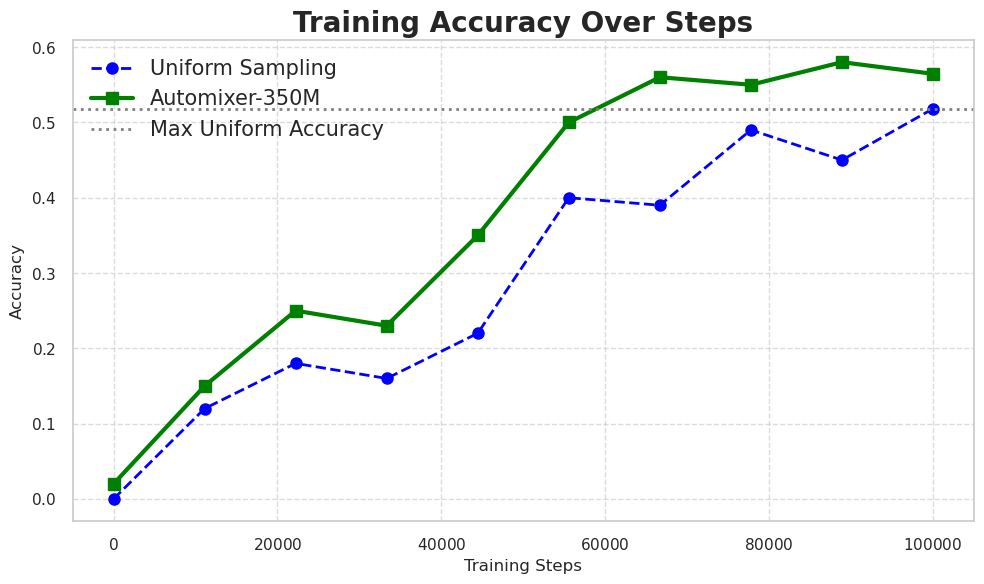}
  \caption{ A performance comparison of two approaches (\emph{uniform sampling} and \emph{AutoMixer-350M}) across ten evenly spaced training steps (0\,--\,100k). Both exhibit minor fluctuations yet follow an overall upward trend in accuracy. \emph{AutoMixer-350M} consistently outperforms \emph{uniform sampling} throughout training, ultimately reaching 56.45\% accuracy versus 51.82\% for \emph{uniform sampling}.}
  \label{fig:trend}
\end{figure}

Despite these occasional variations, the data still validate our two-fold approach. 
First, regrouping data based on capability checkpoints rather than solely relying on fixed domain labels appears crucial for navigating the complexity highlighted in the introduction, where overlapping or ill-defined domains can undermine performance. 
Second, assigning sampling weights to these regrouped sets amplifies the most beneficial samples for each skill, aligning with our objective of pinpointing and upweighting high-value data.


\section{Further Discussion}

\paragraph{Data Regrouping Ablation.}

Table~\ref{tab:checkpoint_sampling} compares three checkpoint-based strategies in terms of their average improvement over a uniform sampling baseline. Relying only on the final (``last'') checkpoint yields a 0.7\% gain, while aggregating all available checkpoints (up to 10) increases that margin to 0.8\%.
The use of selected checkpoints in AutoMixer achieves a more pronounced boost of 1.93\% for 350M scale, demonstrating the advantages of using selected checkpoints to sample data for data regrouping.
Here we show that by leveraging checkpoints that excel in distinct tasks at various stages, and capitalizes on non-monotonic skill emergence, we can more effectively pinpoints those training samples most conducive to each targeted capability.
However, it is also true that earlier Table~\ref{tab:checkpoints} also shows a trend where larger model tends to converge all skills onto one checkpoint, we leave the question of trade-offs between the number of checkpoint samplers and size of model (influence scores' computational speed) for future research.
 

\paragraph{AutoMixer Performance Trajectory.}

Figure~\ref{fig:trend} compares our proposed AutoMixer-350M approach against \emph{uniform sampling} across ten training steps (0-100k). 
The plotted accuracies reveal that AutoMixer-350M not only starts above Uniform but sustains a clear lead throughout training, ultimately reaching 56.45\% accuracy compared to 51.82\% for \emph{uniform sampling}. 
These steady gains underscore AutoMixer’s ability to more effectively reorganize training data over time, allowing the model to focus on samples that yield greater learning benefits. 

\begin{figure}[t]
  \centering
  \includegraphics[width=1.0\columnwidth]{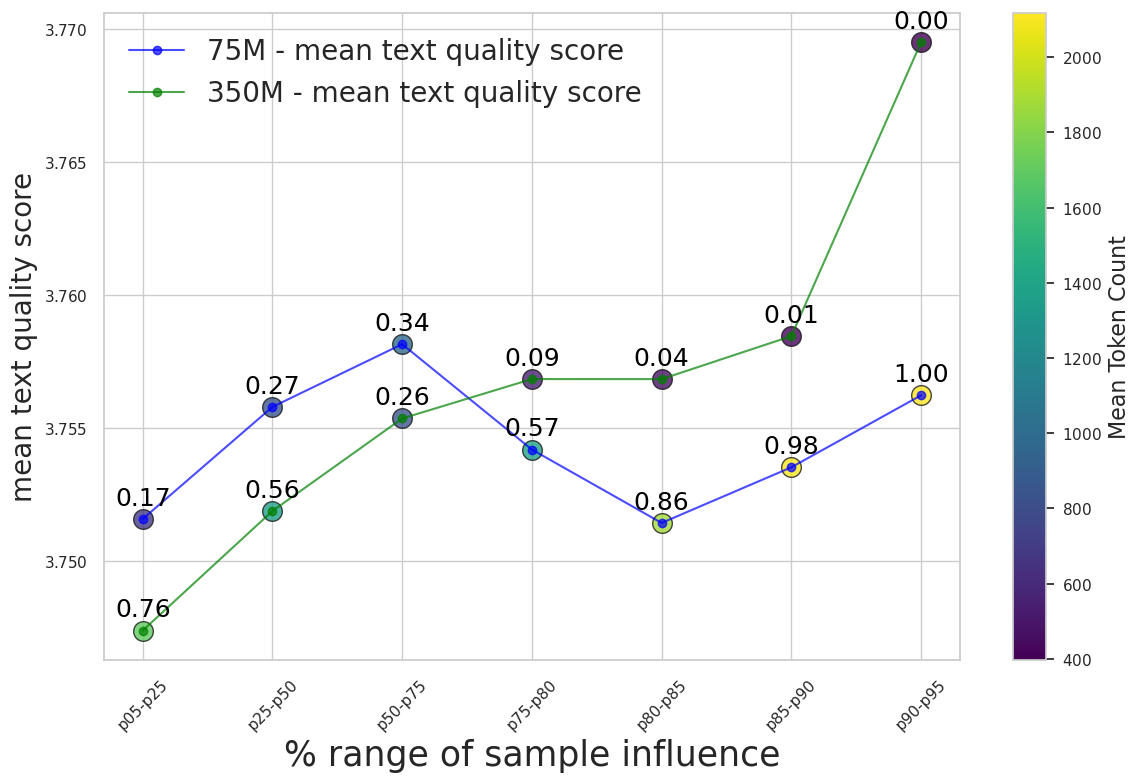}
  \caption{ 
  \textbf{Mean text quality score by range}: 
  We show the sample quality score across all buckets (grouped by percentiles) of samples sorted by influence scores. 
  The \textit{normalized mean token count} (in range $[0,1]$) per sample in the same set of buckets is labeled on each point. 
  75M proxy model tends to select longer sentences with higher influences.
  }
  \label{fig:token}
\end{figure}

\paragraph{Impact of Proxy Model Sizes.}

Figure~\ref{fig:token} reveals an intriguing pattern: smaller proxy models, such as those with 75M parameters, tend to select samples with longer sentences and higher influence scores. 
This observation suggests that the 75M proxy model is particularly adept at identifying influential samples by focusing on longer, more informative sentences. This capability is especially useful in scenarios with limited computational resources, as these models provide valuable signals for sampling task-aware data, even within the lower influence-scoring range.
\begin{figure}[t]
  \centering
  \includegraphics[width=1.0\columnwidth]{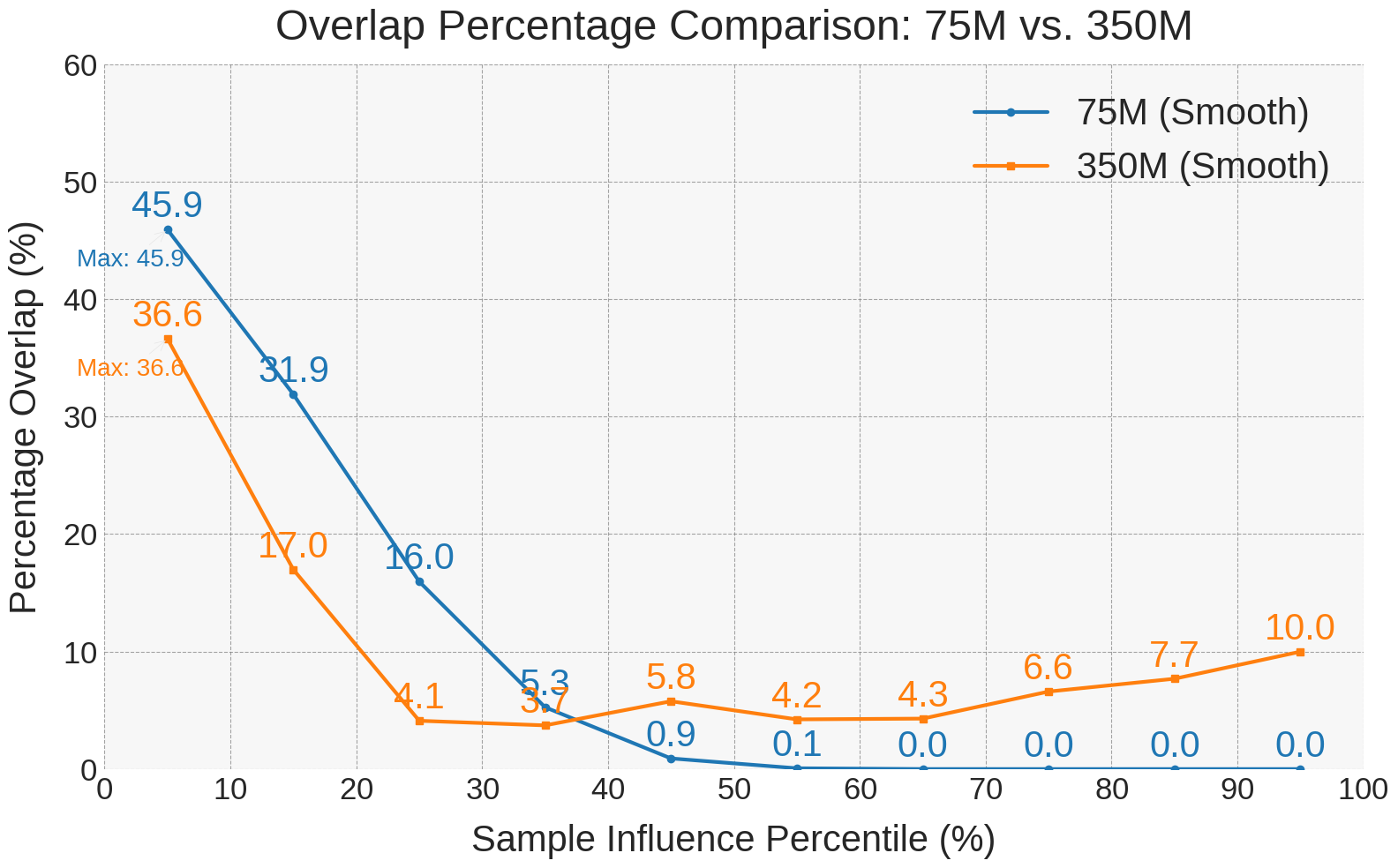}
  \caption{ Plot of sample overlap percentages in respective models (75M \& 350M), measured across increasing influence scores. 
  Two main trends emerge: 
  (1) The 75M model identifies similar set of samples as the larger 350M models for lower-influence samples; while the 350M model captures a higher percentage of higher influence samples and,
  (2) The smaller proxy model’s overlap distribution is notably skewed toward lower sample influences.
  }
  \label{fig:density}
\end{figure}
Conversely, larger proxy models, like the 350M model, excel at distinguishing samples with higher influence scores (See Figure~\ref{fig:density}). 
This implies that increasing model size enhances the ability to discern and prioritize samples that are more impactful for downstream tasks. Larger models are thus better suited for identifying high-quality samples that significantly contribute to the learning process.
These findings highlight two key insights: 
(1) Smaller proxy models can be effectively used to derive useful signals for pruning, especially when focusing on lower influence-scoring samples. 
(2) Larger models offer improved performance in identifying and prioritizing samples with higher influence scores, making them advantageous for tasks requiring high precision in sample selection.

\section*{Conclusion}

Our study demonstrates the effectiveness of \emph{AutoMixer} in enhancing language model pretraining through strategic checkpoint sampling and data regrouping. These results highlight \emph{AutoMixer}'s ability to enhance skill acquisition. 
Moreover, the analysis of proxy model sizes reveals that smaller models, such as those with 75M parameters are suitable for resource-constrained scenarios; 
while larger models, like the 350M model, excel in prioritizing high-impact samples, offering advantages for tasks requiring high precision.
Overall, these findings underscore the potential of optimized data sampling and checkpoint models to significantly boost pretraining performance. 

\section*{Limitations}

Although our multi-checkpoint data mixing strategy demonstrates improvements on diverse reasoning tasks, several constraints remain. 
First, estimating sample influence across multiple checkpoints introduces additional computational overhead, potentially limiting scalability to larger training runs. 
Second, while we highlight how domain overlaps complicate data grouping, our current approach does not explicitly mitigate biases that may arise from imbalanced or skewed datasets. 
Third, the data mixtures and checkpoints used in this work are focused on reasoning benchmarks; applying the same procedure to other domains or more extensive, heterogeneous corpora may reveal different challenges or optimal configurations. 
Finally, as our proxy-based simulations rely on approximate modeling of training dynamics, there is no guarantee that fine-grained influence scores will hold across substantially different architectures or larger-scale training protocols. 
Future work could address these gaps through further studies into how checkpoint selection interacts with diverse model architectures and data domains.

\section*{Ethics Statement}

Our approach leverages public datasets and standard reasoning benchmarks, aiming to refine how training samples are selected and weighted based on evolving model capabilities. 
While this focus can boost efficiency and skill-specific performance, it also raises ethical considerations. 
For instance, optimizing data mixtures for particular tasks may inadvertently deprioritize other skills or amplify existing biases if certain subpopulations are underrepresented or misrepresented in the training data. 
Additionally, the checkpoint-based framework does not inherently account for privacy or fairness concerns, making transparency in data sourcing and audit processes essential. 
Researchers and practitioners employing this method should be mindful of the potential for disproportionate impact on vulnerable groups and consider implementing robust bias detection, inclusive data collection, and clear documentation regarding data provenance. 
As with any technique that refines data selection, careful oversight is necessary to ensure that efforts to enhance performance do not come at the expense of fairness or responsible use.



\bibliography{bibtex/influence,anthology,bibtex/coreset,bibtex/custom,bibtex/datasets,bibtex/nas-overview,bibtex/pretrained,bibtex/zero_nas,bibtex/mllm,bibtex/sampling,bibtex/merging,bibtex/adaptation,inf_literature/ref}
\bibliographystyle{acl_natbib}

\clearpage

\appendix

\onecolumn

\appendix

\section{Dataset Analysis}

\begin{table}[h]
\centering
\resizebox{\textwidth}{!}{%
\begin{tabular}{lrrrrr}
\toprule
Range & Mean Language Score & Mean Score & Mean Int Score & Mean Token Count & Sum Token Count \\
\midrule
p80-p85 & 0.97 & 3.76 & 4.01 & 474.01 & 1,593,100,685 \\
p85-p90 & 0.97 & 3.76 & 4.01 & 413.55 & 1,388,916,981 \\
p75-p80 & 0.97 & 3.76 & 4.00 & 550.58 & 1,852,164,429 \\
p05-p25 & 0.96 & 3.75 & 4.00 & 1,698.47 & 22,912,588,553 \\
p90-p95 & 0.97 & 3.77 & 4.01 & 397.05 & 1,345,066,462 \\
p25-p50 & 0.96 & 3.75 & 4.00 & 1,353.91 & 22,789,192,924 \\
p50-p75 & 0.96 & 3.76 & 4.00 & 846.16 & 14,215,340,759 \\
\bottomrule
\end{tabular}%
}
\caption{Summary statistics for different percentile ranges for AutoMixer-350M.}
\label{tab:summary_statistics}
\end{table}

\section{Data Regrouping}

In our experiments, we utilized a data regrouping strategy to enhance the quality and relevance of our dataset.

For each of the $K$ checkpoints, we performed the following operations:

\begin{enumerate}
    \item \textbf{Percentage Calculation}: We calculated the approximate percentiles of the \texttt{influence} metric within our dataset.

    \item \textbf{Table Creation}: We created a new table to store data with influence values above a certain threshold.

    \item \textbf{Influence Aggregation}: We aggregated influence scores across multiple tables to create a new column \texttt{total\_influence}.
\end{enumerate}

\paragraph{Weight Estimation.}

To estimate the weight of each table, we calculated the scaled influence and the total influence token product. This involved computing the minimum and maximum influence values and scaling the influence accordingly.

\begin{algorithm}
\caption{Data Processing Pipeline}
\begin{algorithmic}[1]
\STATE \textbf{Input:} Dataset $D$, Threshold $T$
\STATE \textbf{Output:} Total Influence Token Product
\STATE
\STATE \textbf{Procedure} \textsc{DataRegrouping}
    \STATE Calculate percentiles of influence in $D$
    \STATE Create a filtered dataset $F$ where influence $\geq T$
\STATE \textbf{End Procedure}
\STATE
\STATE \textbf{Procedure} \textsc{InfluenceAggregation}
    \STATE Aggregate influence scores in $F$ to compute total influence for each ID
    \STATE Store results in a new table $A$
\STATE \textbf{End Procedure}
\STATE
\STATE \textbf{Procedure} \textsc{WeightEstimation}
    \STATE Calculate min and max influence from $A$
    \STATE Compute scaled influence for each entry in $A$
    \STATE Calculate total influence token product
\STATE \textbf{End Procedure}
\STATE
\STATE \textbf{Return} Total Influence Token Product
\end{algorithmic}
\end{algorithm}

\end{document}